\documentclass{article}
\usepackage{spconf,amsmath,graphicx}

\usepackage{enumitem}
\usepackage{url}
\setlist{nosep, leftmargin=14pt}

\usepackage{mwe} 

\title{LUMEN: Longitudinal Multi-Modal Radiology Model for Prognosis and Diagnosis}
\name{
\begin{tabular}{c}
Zhifan Jiang$^1$ \qquad
Dong Yang$^2$  \qquad
Vishwesh Nath$^2$  \qquad
Abhijeet Parida$^{1,3}$  \qquad \\
Nishad P. Kulkarni$^1$  \qquad 
Ziyue Xu$^2$  \qquad 
Daguang Xu$^2$  \qquad 
Syed Muhammad Anwar$^{1,4}$  \qquad \\ 
Holger R. Roth$^2$  \qquad 
Marius George Linguraru$^{1,4}$
\end{tabular}
}
\address{
$^1$ Sheikh Zayed Institute for Pediatric Surgical Innovation, \\ 
Children's National Hospital, Washington DC, USA \\
$^2$ Nvidia Corporation, Santa Clara, CA, USA \\
$^3$ ETSI Telecomunicaci\'{o}n, 
Universidad Polit\'{e}cnica de Madrid, Madrid, Spain \\
$^4$ School of Medicine and Health Sciences, George Washington University, Washington DC, USA 
}
\begin{document}
%
\maketitle
\begin{abstract}
Large vision-language models (VLMs) have evolved from general-purpose applications to specialized 
use cases such as in the clinical domain, demonstrating potential for decision support in radiology. 
One promising application is assisting radiologists in decision-making by the analysis of radiology 
imaging data such as chest X-rays (CXR) via a visual and natural language question-answering (VQA) 
interface. When longitudinal imaging is available, radiologists analyze temporal changes, which are 
essential for accurate diagnosis and prognosis. The manual longitudinal analysis is a time-consuming 
process, motivating the development of a training framework that can provide prognostic 
capabilities. We introduce a novel training framework LUMEN that is optimized for longitudinal CXR 
interpretation, leveraging multi-image and multi-task instruction fine-tuning to enhance prognostic 
and diagnostic performance. We conduct experiments on the publicly available MIMIC-CXR and its 
associated Medical-Diff-VQA datasets. We further formulate and construct a novel instruction-following 
dataset incorporating longitudinal studies, enabling the development of a prognostic VQA task. Our 
method demonstrates significant improvements over baseline models 
in 
diagnostic VQA tasks, and more importantly, shows promising potential for prognostic capabilities. 
These results underscore the value of well-designed, instruction-tuned VLMs in enabling more 
accurate and clinically meaningful radiological interpretation of longitudinal radiological imaging data.

\end{abstract}
\section{Introduction}
\label{sec:intro}

The workload in radiology departments has been steadily increasing~\cite{bruls2020-s}. 
Recently, large vision-language models (VLMs) have advanced rapidly, extending from general-purpose applications to clinical domains. VLMs present a promising tool for tackling complex and demanding medical tasks, including radiologic interpretation and decision support.

LLaVA~\cite{llava2023-s} introduced the novel learning paradigm of instruction fine-tuning for pre-trained large models in the multi-modal domain, achieving strong zero-shot performance on vision-language tasks. Since the introduction of LLaVA, multiple open-source multi-modal models have emerged, including VILA~\cite{vila2024cvpr-s} and its updated version NVILA~\cite{liu2024nvila}. DeepSeek-VL2~\cite{deepseek-vl2}, Qwen2-VL~\cite{qwen2-vl}, MiniGPT-V2~\cite{minigpt-v2} are additional examples of open-access VLMs. LLaVA and VILA models stand out as the most accessible for user-driven fine-tuning, allowing broader customization. Both have been adapted towards biomedical applications, leading to the development of specialized models such as LLaVA-Med~\cite{li2023llava-med-s} and VILA-M3~\cite{nath2024vilam3}, designed for medical vision-language tasks.

In the domain of artificial intelligence (AI)-based radiographic understanding, the publicly available MIMIC-CXR dataset~\cite{mimic3,mimic1,mimic2} and its derived Medical-Diff-VQA dataset~\cite{mimicvqa1,mimicvqa2} serve as valuable resources for evaluating diagnostic vision question answering (VQA). These datasets have become benchmarks for many related studies, providing structured QA pairs that facilitate the development and evaluation of VLMs in radiology. Among these, the D-Rax model~\cite{nisar2025drax-s} introduced a novel approach by incorporating expert model predictions from state-of-the-art (SOTA) classifiers into instruction-tuning data. This enhancement proposed by D-Rax improved the model's capabilities in abnormality detection. VILA-M3 further extended the role of medical expert knowledge and also incorporated segmentation data alongside diagnostic labels. However, current VLMs in radiology are typically limited to single-image understanding and cannot perform the temporal reasoning that radiologists routinely do. In practice, radiologists compare current and prior studies to assess disease progression or treatment response. Yet, most published medical VLMs do not incorporate longitudinal information. The emerging differential VQA task highlights this gap. Only recently have a few works attempted such sequential image reasoning~\cite{mimicvqa2,cho2024pretraining,lu-miccai2024-diff-s,yung-miccai2024-diffvlm}. These models have focused on comparing a main image with a reference image  extracted from two radiological studies at different time points, aiming to enhance the model's capabilities in detecting differences and assessing disease progression.

Beyond diagnosis and description, radiologists often prognosticate – predicting patient outcomes based on imaging findings. Traditional deep learning approaches address these questions using purely image-based models or simple combinations of image features with clinical data~\cite{jiao2021covid,miccai2024covid}. However, no current VLMs have been developed specifically for prognosis. Developing a prognostic VLM requires the ability to learn from longitudinal data, reason over temporal changes, and handle outcome-related questions, many of which lack a definitive ground-truth answer.

In this work, we present LUMEN, a unified VLM designed to address these limitations. Built on NVILA, our model efficiently processes single or sequential input images, enabling end-to-end VQA for both diagnostic and prognostic tasks. We integrate expert model predictions and time intervals from DICOM.
A key challenge in prognosis is the lack of reference instruction-following data. To overcome this, we generated an in-house prognostic instruction dataset based on Medical-Diff-VQA. By leveraging known temporal changes between studies extracted from the difference QA pairs, we formulated prognosis-related questions and corresponding outcome predictions. Our \textbf{main contributions} are: (1) Enhancing Medical-Diff-VQA with longer, large language model (LLM)-generated responses for more informative interactions beyond short factual answers; (2) Creating prognostic instruction-following data, incorporating diverse outcome-based questions and predicted answers; (3) Developing a unified VLM capable of processing multiple images, supporting both diagnostic interpretation and prognostic forecasting.

\section{Methods}
\label{sec:method}

\subsection{Data}
\label{subsec:data}

We utilized the MIMIC-CXR dataset~\cite{mimic3,mimic1,mimic2} and an updated version of the Medical-Diff-VQA dataset~\cite{mimicvqa1,mimicvqa2}. MIMIC-CXR is a large publicly available dataset of 377,110 CXRs from 227,827 studies, with structured labels extracted from free-text radiology reports.  
The Medical-Diff-VQA dataset is derived from MIMIC-CXR and comprises 700,703 QA pairs. A single image is extracted from a study. The questions span seven categories: abnormality, presence, view, location, level, type, and difference. A difference question involves comparisons between a pair of main and reference images to assess changes over time. QA pairs have two types: 1) open-ended questions (\textit{e.g.}, "what abnormalities are seen in this image?"), which expect descriptive, variable-length natural language responses; 2) close-ended questions (\textit{e.g.}, "is there evidence of any abnormalities in this image?"), typically answered in a binary "yes/no". 

We considered the first six question categories (abnormality, etc.) as diagnostic questions requiring only a single input image. We followed the official Medical-Diff-VQA dataset partitioning and extracted one study per subject to form the test set. This resulted in 1) a training set: 129,231 images, 428,995 diagnostic questions (51\% open-ended and 49\% close-ended), and 43,381 difference questions; 2) a test set: 4,190 images, 13,688 diagnostic questions (49\% open-ended and 51\% close-ended), and 1,369 difference questions.


\subsection{Instruction Enhancement}
\label{subsec:ens}
To enhance the naturalness of the instructions and improve the readability, we refined the answers within the Medical-Diff-VQA dataset. While the original factual instructions were easy for evaluation, it lacked the fluency and readability needed for real-world clinical applications. We used the Llama-3.2-11B-Vision-Instruct model~\cite{llama3} to expand each answer into a complete sentence, while maintaining its original meaning with an engineered prompt.

\subsection{Expert Model Predictions}
We followed the strategy outlined in~\cite{nisar2025drax-s} to improve the instruction-following data from Medical-Diff-VQA by adding expert model predictions generated using TorchXRayVision~\cite{torchxray}. These SOTA expert models provided predictions for disease diagnosis (18 conditions), patient age, race (Asian, Black, White), and image view (frontal, lateral) for each image in the dataset. To enhance interpretability for the LLM, disease probabilities were mapped to a text-based format with four confidence levels: no, maybe no, maybe yes, and yes, based on calibrated probability thresholds at 0.25, 0.5, 0.75.

\subsection{Longitudinal Instructions Design}

To extend the capabilities of the dataset to prognosis beyond retrospective comparisons, we generated future prediction questions using factual information extracted from difference questions within Medical-Diff-VQA. The original set included questions comparing two images of the same patient taken at different time points. To formulate prognostic questions, we used Llama-3.2-11B-Vision-Instruct~\cite{llama3} with a structured prompt designed to predict future changes in the reference image. For each QA pair, we generated an additional answer to a randomly selected question from a predefined set of 100 questions, \textit{e.g.}, “What changes can be expected in this patient’s chest X-ray in $<$placeholder$>$ days?” The selected question, along with the difference answers, was then processed using an engineered prompt. The constructed instruction-tuning dataset will be made publicly available for reproducible research, pending PhysioNet approval.

\subsection{Experiments}
\noindent\textbf{Training}
We fine-tuned the pretrained NVILA-8B on the MIMIC-CXR dataset using the training and test sets defined in Section~\ref{subsec:data} for both diagnostic and prognostic tasks. Models were trained using the enriched dataset with longer, LLM-generated responses. Specifically, we fine-tuned NVILA-8B using diagnostic instructions with single image input and using both diagnostic and enhanced prognostic instructions with two-image input (LUMEN).
The projection layer, the language model, and the vision encoder are updated. Training was conducted for one epoch, with a learning rate of $1.5e^{-5}$  and a global batch size of $128$, utilizing four NVIDIA H100 GPUs (80GB memory each). 

\noindent\textbf{Evaluation Metrics}
Performance was first evaluated using BLEU-4 and ROUGE-L scores to access lexical similarity. Additionally, following~\cite{li2023llava-med-s,nisar2025drax-s}, we used short answers in the original Medical-Diff-VQA to compute the token recall, measuring the ratio of correctly generated tokens relative to the reference tokens for open-ended questions. For close-ended questions, accuracy was used as the evaluation metric. 

Instruction fine-tuning on the more comprehensive dataset, generated by an LLM to include longer and more descriptive responses, can help the model learn both to provide factual answers and to elaborate when required. 
However, these conventional scores may not be robust in this setting. BLEU and ROUGE often rated outputs highly even when they contained clinically incorrect information due to surface-level lexical overlap. Relying solely on n-gram metrics can be inadequate for medical text evaluation~\cite{liu-miccai2024-mrscore-s}.

To address this limitation, we incorporated Llama score, an LLM-based metric computed by prompting Llama-3.1-405B to directly compare generated responses with the reference on a scale of 1 to 10. This approach provides a more nuanced assessment of clinical correctness and helpfulness. Notably, cases where traditional scores were high (\textit{e.g.}, ROUGE-L=0.7) but the generated response was incorrect (\textit{e.g.}, a model responding "No evidence of any abnormalities is present in this image" when the reference was "Yes, there is evidence of abnormalities present in this image") were appropriately penalized by Llama score (\textit{e.g.}, scoring only 1/10). This demonstrates that the Llama Score is a more reliable metric in radiological VQA tasks, where factual correctness outweighs lexical similarity.

\section{Results and Discussion}
\begin{figure*}
    \centering
    \includegraphics[width=\textwidth]{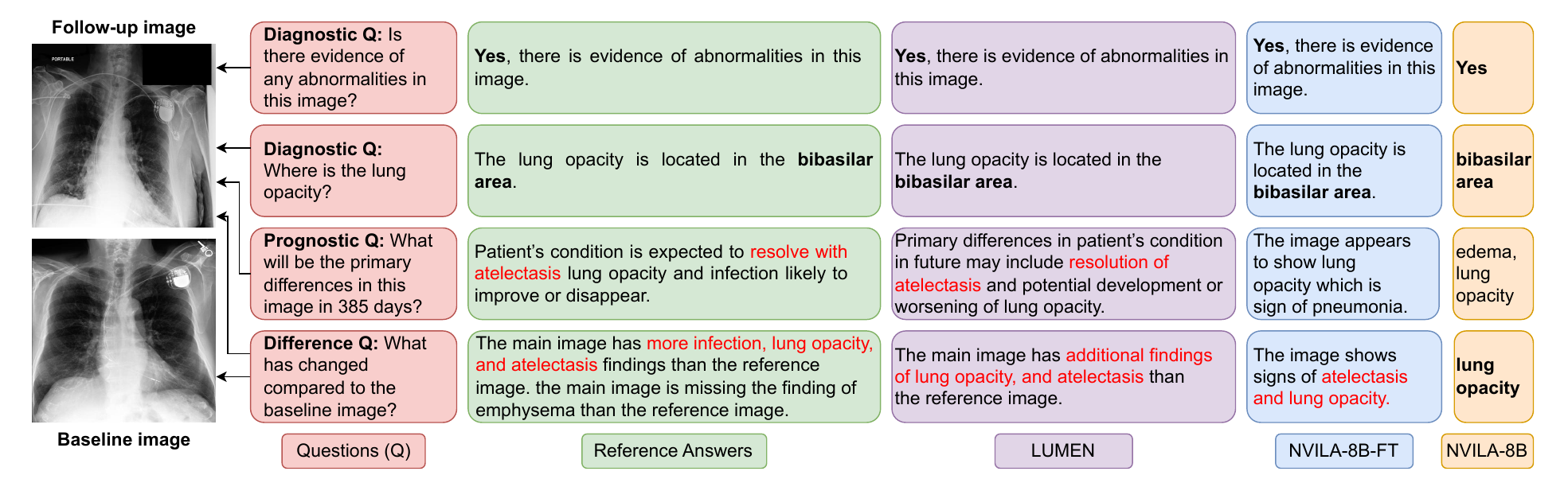}
    \caption{Qualitative evaluation: conversations provided by NVILA-8B, NVILA-8B-FT (finetuned), and LUMEN.} 
\label{fig2}
\end{figure*}

\subsection{Prognostic (Temporal) Question Answering}
As shown in Table~\ref{tab:res_diff}, the models showed significantly greater difficulty when answering difference-based and prediction-oriented questions, which required temporal reasoning. NVILA-8B, fine-tuned only on diagnostic tasks, performed poorly when prompted with difference-based questions comparing two time-separated images. However, LUMEN explicitly trained with temporal information improved significantly, suggesting that explicit exposure to temporal image pairs helped the model capture disease progression patterns. Despite these gains, overall performance remained weaker than in diagnostic tasks, as difference-based reasoning requires both precise lesion localization and an understanding of disease evolution over time. Existing studies~\cite{cho2024pretraining,lu-miccai2024-diff-s} solely focus on the difference-based reasoning can be incorporated to enhance our model's ability to analyze longitudinal changes. Qualitative results were presented in Figure~\ref{fig2}.

\begin{table}[thpb]
\caption{Prognostic Performance: evaluated on open-ended difference- and prediction-based questions. ${\dagger}$ = fine-tuned on Medical-Diff-VQA. The asterisks show
statistical significance (p $< 0.001$) across paired comparisons with LUMEN, using the Wilcoxon signed rank test.}
\label{tab:res_diff}
\centering
\begin{tabular}{llll}
\hline
Model & BLEU-4 & ROUGE-L & Llama Score \\
\hline
\multicolumn{4}{c}{Test on $1,369$ images and $1,369$ \textbf{difference} questions} \\
NVILA-8B & $0.001$* & $0.071$* & $2.486$* \\
NVILA-8B$^{\dagger}$ & $0.020$* & $0.206$* & $3.275$* \\
LUMEN$^{\dagger}$ & $\mathbf{0.375}$ & $\mathbf{0.656}$ & $\mathbf{4.611}$ \\
\hline
\multicolumn{4}{c}{Test on $2,451$ images and $3,936$ \textbf{prediction} questions} \\
NVILA-8B & $0.003$* & $0.052$* & $2.242$* \\
NVILA-8B$^{\dagger}$ & $0.019$* & $0.167$* & $3.177$* \\
LUMEN$^{\dagger}$  & $\mathbf{0.095}$ & $\mathbf{0.303}$ & $\mathbf{4.866}$ \\
\hline
\end{tabular}
\end{table}

\subsection{Diagnostic Question Answering}
As shown in Table~\ref{tab:res_open}, instruction fine-tuning the NVILA-8B model on a specific clinical QA dataset dramatically improved the performance on diagnostic questions. BLEU, ROUGE, and Llama scores were not computed for finetuned LLaVA-Med and D-Rax, as they generated only short answers consistent with the original Medical-Diff-VQA. This heavily penalized lexical metrics, as observed with NVILA-8B, which also produces simple responses. Instead, recall and accuracy from the original paper~\cite{nisar2025drax-s} were used for comparison.
For close-ended questions, the fine-tuned model achieved near-perfect lexical overlap with reference answers, reflected in high BLEU-4 and ROUGE-L scores and a notable accuracy increase. the fine-tuned model’s open-ended question answers also improved substantially, though performance is inherently lower than close-ended cases. Llama score also improved after fine-tuning, mirroring the trend in BLEU and ROUGE scores. This alignment suggests that the fine-tuned model’s answers are not just lexically closer to the references, but also more clinically sound and complete.

Notably, incorporating multi-task instruction training (including both diagnostic and temporal tasks) did not degrade diagnostic performance – the multi-task model (LUMEN) maintained comparable accuracy and language scores. This consistency demonstrates that the model can learn diverse question types simultaneously, preserving its diagnostic interpretative ability even when trained jointly with more complex prognostic tasks.

\begin{table}[thpb]
\caption{Diagnostic Performance: evaluated on questions related to abnormality, disease presence, location, severity level, type, and image view. ${\dagger}$ = fine-tuned on Medical-Diff-VQA. BLEU-4 and ROUGE-L were used. The asterisks show
statistical significance (p $< 0.05$) across paired comparisons with LUMEN, using the Wilcoxon signed rank test.}
\label{tab:res_open}
\centering
\begin{tabular}{lllll}
\hline
Model & BLEU & ROUGE & Recall & Llama \\
\hline
\multicolumn{5}{c}{Test on $3,228$ images and $6,683$ \textbf{open-ended} questions} \\
LLaVA-Med$^{\dagger}$ & & &  $0.604$ & \\
D-Rax$^{\dagger}$ & & &  $0.616$ & \\
NVILA-8B & $0.025$* & $0.258$ & $0.674$ & $6.975$* \\
NVILA-8B$^{\dagger}$ & $0.551$ & $0.766$ & $0.689$* & $7.933$ \\
LUMEN$^{\dagger}$ & $0.551$ & $0.766$  & $\mathbf{0.694}$ & $\mathbf{7.953}$ \\
\hline
Model & BLEU & ROUGE & Accuracy & Llama \\
\hline
\multicolumn{5}{c}{Test on $4,157$ images and $7,005$ \textbf{close-ended} questions} \\
LLaVA-Med$^{\dagger}$ & &  & $0.77$ & \\
D-Rax$^{\dagger}$ & & & $0.79$ & \\
NVILA-8B & $0.014$* & $0.182$* & $0.744$* & $6.047$* \\
NVILA-8B$^{\dagger}$ & $0.775$* & $0.900$ & $\mathbf{0.865}$ & $\mathbf{8.583}$ \\
LUMEN$^{\dagger}$ & $0.772$ & $0.898$ & $0.863$ & $8.565$ \\
\hline
\end{tabular}
\end{table}

Prediction-oriented questions proved to be the most challenging. Even with task-specific fine-tuning, models struggled to predict the progression of findings. Notably, models trained on only the original MIMIC-CXR dataset performed the worst in this task, as they had no exposure to future temporal dependencies. On the other hand, our new model demonstrated improvements indicating some capacity to generalize patterns in disease progression. However, forecasting patient outcomes from reference images remains an open challenge. The difficulty arises from the inherent uncertainty in disease trajectories, received treatment, and the lack of explicit longitudinal ground truth in the dataset. Another limitation of our model is that it incorporates only two images corresponding to two studies at different time points. Using a sequence of multiple temporal images could further enhance the benefits of longitudinal data training. 

\section{Conclusion}
We developed LUMEN, an all-in-one VLM, that incorporates longitudinal data and is capable of both clinical diagnosis and prognosis. The NVILA-8B base model was fine-tuned on the MIMIC-CXR dataset with specialized reasoning instructions. Our findings illustrate both the promise and limitations of large VLMs in a clinical setting. Models trained with temporal information show potential in tracking disease progression. Future work should explore multi-modal training that incorporates comprehensive clinical information such as treatment to further enhance clinical validity.

\section{Compliance with ethical standards}
\label{sec:ethics}

This research study was conducted retrospectively using human subject data made available 
in open access by A. Johnson \textit{et al.} (2024). MIMIC-IV (version 3.1). PhysioNet. RRID:SCR\_007345. \url{doi.org/10.13026/kpb9-mt58}. 
The collection of patient information and creation of the research resource was reviewed by the Institutional Review Board at the Beth Israel Deaconess Medical Center, who granted a waiver of informed consent and approved the data sharing initiative.

\section{Acknowledgments}
\label{sec:acknowledgments}

This work was supported by The National Cancer Institute (UG3CA236536).

\bibliographystyle{IEEEbib}
\bibliography{refs}

\end{document}